\documentclass{article}

\usepackage{microtype}
\usepackage{graphicx}
\usepackage{subfigure}
\usepackage{booktabs} 
\usepackage{multirow}

\usepackage{hyperref}

\usepackage[accepted]{icml2023}

\usepackage{amsmath}
\usepackage{amssymb}
\usepackage{mathtools}
\usepackage{amsthm}

\usepackage[capitalize,noabbrev]{cleveref}

\theoremstyle{plain}

\theoremstyle{definition}

\theoremstyle{remark}

\usepackage[textsize=tiny]{todonotes}

\icmltitlerunning{Fed-CPrompt: Contrastive Prompt for Rehearsal-Free Federated Continual Learning}

\begin{document}

\twocolumn[
\icmltitle{Fed-CPrompt: Contrastive Prompt for Rehearsal-Free Federated Continual Learning}



\icmlsetsymbol{equal}{*}

\begin{icmlauthorlist}
\icmlauthor{Gaurav Bagwe}{ecemtu}
\icmlauthor{Xiaoyong Yuan}{compmtu}
\icmlauthor{Miao Pan}{eceh}
\icmlauthor{Lan Zhang}{ecemtu}
\end{icmlauthorlist}

\icmlaffiliation{ecemtu}{Department of ECE, Michigan Technological University, Houghton, MI, USA}
\icmlaffiliation{compmtu}{College of Computing, Michigan Technological University, Houghton, MI, USA}
\icmlaffiliation{eceh}{Department of ECE, University of Houston, Houston, TX, USA}

\icmlcorrespondingauthor{Gaurav Bagwe}{grbagwe@mtu.edu}

\icmlkeywords{Federated Continual Learning, Federated learning, Rehearsal-free Continual Learning, Prompt Learning}

\vskip 0.3in
]



\printAffiliationsAndNotice{}  

\begin{abstract}
Federated continual learning (FCL) learns incremental tasks over time from confidential datasets distributed across clients. This paper focuses on rehearsal-free FCL, which has severe forgetting issues when learning new tasks due to the lack of access to historical task data. To address this issue, we propose Fed-CPrompt based on prompt learning techniques to obtain task-specific prompts in a communication-efficient way. Fed-CPrompt introduces two key components, asynchronous prompt learning, and contrastive continual loss, to handle asynchronous task arrival and heterogeneous data distributions in FCL, respectively. Extensive experiments demonstrate the effectiveness of Fed-CPrompt in achieving SOTA rehearsal-free FCL performance.

\end{abstract}

\section{Introduction}

Federated learning (FL) has been a popular collaborative machine learning paradigm enabling multiple clients to learn a shared model without exposing private client data~\cite{mcmahan2017communication}. While successful, existing FL algorithms are mainly designed for a single task with fixed datasets on clients~\cite{mcmahan2017communication,li2020federated,li2021model}, which becomes ineffective in handling non-stationary data distribution over time. Therefore, recent efforts have been put into federated continual learning (FCL) to learn tasks that are presented sequentially. Since the model in continual learning (CL) may overfit data from the current task and suffer from catastrophic forgetting~\cite{kirkpatrick2017overcoming}, the mainstream research to address the forgetting issue can be roughly divided into two categories: rehearsal-based and rehearsal-free FCL.

Although rehearsal-based approaches achieve state-of-the-art (SOTA) performance by using the rehearsal buffer to store and retrain data from previous tasks, the buffer size needs to be large enough to effectively mitigate forgetting \cite{wang2022learning,dong2022federated}, leading to scalability and data storage constraints in FL. Moreover, many applications do not allow this buffer due to privacy concerns~\cite{smith2022closer}, further restricting their adoption in practice. Hence, this work focuses on rehearsal-free FCL. Existing efforts along this line regularize the global model with knowledge from previous tasks when learning a new task~\cite{shoham2019overcoming,yoon2021federated,casado2022concept,usmanova2021distillation,usmanova2022federated,ma2022continual}. Unfortunately, they have substantially deteriorated performance compared to rehearsal-based approaches~\cite{wang2022learning}. Moreover, existing research requires continuously exchanging the entire model to learn incremental tasks in FCL, leading to significant communication overhead.  In view of these, it is critical to developing \textit{innovative rehearsal-free FCL in a communication-efficient way to address the forgetting issue while maintaining the  model plasticity for new tasks}.

Enlightened by the recent advance of prompting techniques \cite{lester2021power,liu2022p,li2021prefix}, in this work, we leverage prompt learning to achieve the above goal. As one promising transfer learning approach, prompt learning uses insertable embeddings called prompts to condition a pre-trained model for downstream tasks. Recent research enables prompt-based CL by using key-query mechanisms, which achieves SOTA rehearsal-free performance, even outperforming rehearsal-based CL~\cite{wang2022learning,wang2022dualprompt,smith2022coda}. Due to the small size of prompt parameters, the communication efficiency of FCL is expected to be improved significantly. However, existing prompt-based CL is designed for centralized datasets, which becomes ineffective in FL with distributed and confidential datasets. The main limitation is due to the inherent heterogeneity of distributed clients. On the one hand, clients may observe heterogeneous data for the same task, leading to biased learning performance and slow convergence. On the other hand, incremental tasks may arrive asynchronously on clients, further deteriorating the overall learning performance. Therefore, to unleash the potential of prompting for rehearsal-free FCL, we propose Fed-CPrompt to facilitate inter-task and inter-client prompt-based knowledge transfer while addressing the heterogeneity concerns of data distribution and task arrival over clients.
Our key contributions are summarized below:
\begin{itemize}
    \vspace{-1em}
    \item We propose Fed-CPrompt, an innovative rehearsal-free FCL framework based on prompting techniques. Fed-CPrompt achieves SOTA FCL performance to handle the stability-plasticity dilemma under heterogeneous FL environments in a communication-efficient way.
    \vspace{-0.5em}\item We introduce two key components to Fed-CPrompt: \textit{asynchronous prompt learning} takes advantage of task asynchronicity to strengthen the task-specific prompts; \textit{C2L loss} alleviates inter-task forgetting and inter-client data heterogeneity via a contrastive and continual loss. 
    \vspace{-0.7em}\item We conduct extensive experiments to demonstrate the effectiveness of Fed-CPrompt in various challenging FCL settings, such as heterogeneous data distribution and asynchronous task arrival.  
\end{itemize}



\begin{figure*}[tbh]
    \centering
    \includegraphics[width=0.8\textwidth]{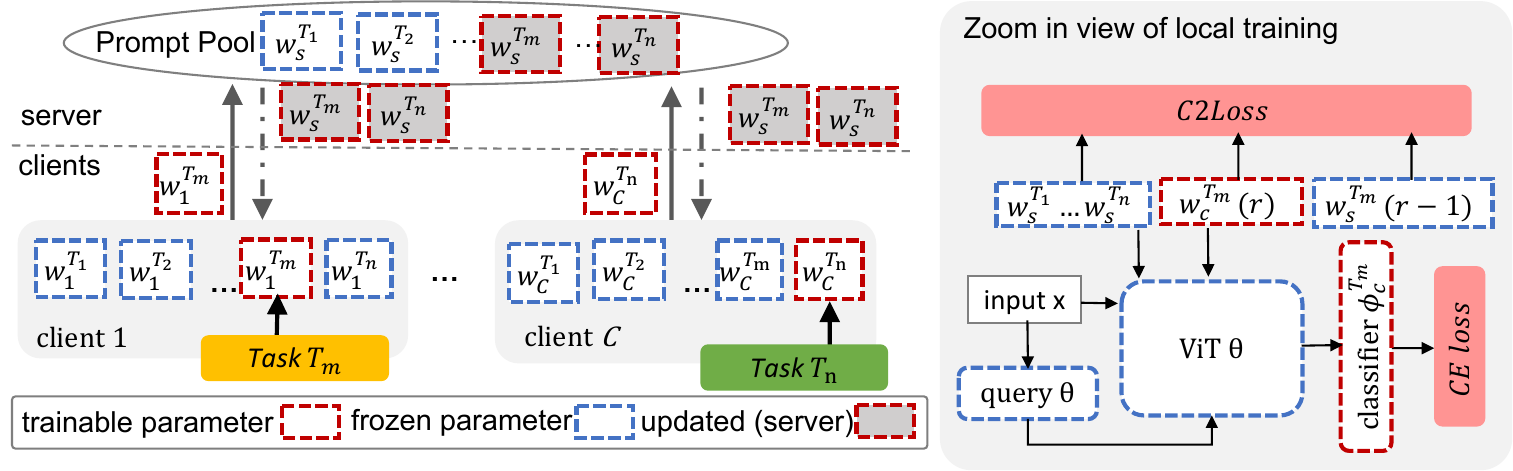}
    \caption{Overview of Fed-CPrompt. {Left}: overall system framework. A server maintains a pool of task-specific prompts to coordinate multiple clients for FCL. The clients may observe different tasks at the same time. Fed-CPrompt allows asynchronous prompt learning. Right: zoom in view of client $c$'s local training for task $\mathcal{T}_m$. Since only the trainable parameters ($w_c^{\mathcal{T}_m}$ and $\phi_c^{\mathcal{T}_m}$) need to be exchanged, Fed-CPrompt is a communication-efficient FCL approach.  
    }
    \label{fig:Framework_Overview}
\end{figure*}


\section{Proposed Method} 
\subsection{Problem Statement}
In a standard FCL setting, a central server coordinates a set of distributed clients $\mathcal{C}$ to learn incremental tasks ${\mathcal{T}_1, \ldots, \mathcal{T}_n}$ over time. 
The training data for each task is distributed to clients and cannot be shared. FCL aims to obtain a global model parameterized by $\mathbf{w}$ to perform all existing tasks. In this work, we consider a challenging CL problem, class-incremental CL, where the task labels are unknown during inference~\cite{dong2022federated}. Our design can be easily extended to the task- or domain-incremental FCL problems. The optimization objective can be written as 
\begin{equation}\label{eq:FCL}
    \min_{\mathbf{w}}  \sum_{i\in\{1, \ldots,n\}} \sum_{c\in\mathcal{C}} \frac{\mathrm{n}_c^{\mathcal{T}_i}}{\mathrm{n}^{\mathcal{T}_i}} \mathcal{L}(\mathcal{D}_c^{\mathcal{T}_i};\mathbf{w}),
\end{equation}
where $\mathrm{n}_c^{\mathcal{T}_i}$ and $\mathrm{n}^{\mathcal{T}_i}$ represent the number of training samples from client $c$ and all clients for task $\mathcal{T}_i$, respectively. $\mathcal{D}_c^{\mathcal{T}_i}$ is the training dataset of $\mathcal{T}_i$ on client $c$.

This objective function uses data from all existing tasks, making it a rehearsal-based FCL problem.
This work focuses on rehearsal-free FCL. Specifically, each client can only observe the training data of the current task, i.e., when training on task $\mathcal{T}_n$, training data of all previous tasks are unseen.
However, due to the unavailability of historical task data, training the current task can overwrite previous task information of the model $\mathbf{w}$ in (\ref{eq:FCL}), deteriorating the forgetting issues in CL. Thus, existing rehearsal-free FCL approaches cannot achieve comparable performance to rehearsal-based approaches~\cite{wang2023federated}.

\subsection{Design Principle}
In this work, we aim to accommodate the forgetting issue for rehearsal-free FCL. Inspired by the success of the prompt-based rehearsal-free CL that achieves SOTA performance, we intend to implement prompting techniques in our design.  Existing prompt-based CL~\cite{smith2022coda, wang2022learning,wang2022dualprompt} use insertable embeddings, called prompts $p$, to condition a frozen pre-trained model $\theta$ to perform incremental tasks. Due to the small size of prompt parameters, a task-specific prompt is created and stored for each task to avoid overwriting previous knowledge. Here, we refer readers to Appendix~\ref{sec:preliminaries} for more details. While successful, the above prompt-based CL research is designed for centralized datasets, which becomes ineffective in FL settings. 

The main challenge of implementing prompting techniques in FCL is the inherent heterogeneity of distributed clients. On the one hand, the data heterogeneity among clients leads to biased local updates and slow convergence. On the other hand, the sequential tasks may appear asynchronously over clients, further delaying convergence.  Due to the small size of learnable parameters in prompt learning, it is essential to improve their learning capacity by facilitating knowledge transfer between tasks and clients. Therefore, we propose Fed-CPrompt, an innovative prompt-based rehearsal-free FCL framework. As shown in Figure~\ref{fig:Framework_Overview},  Fed-CPrompt introduces two key components, asynchronous prompt learning and contrastive and continual loss, to address the aforementioned task arrival and data heterogeneity concerns. In the following, we first introduce these two components and then present the overall training of Fed-CPrompt.

\subsection{Asynchronous Prompt Learning}\label{sec: APL}
We adopt the existing prompt-based CL approach (CODA-P~\cite{smith2022coda}) on clients to learn incremental tasks based on their local data. In CODA-P, the prompt for the current task is re-weighted based on previous task information to refine task-specific representation via attention mechanisms (see Appendix~\ref{sec:preliminaries}). In Fed-CPrompt, when client $c\in\mathcal{C}$ learns task $\mathcal{T}_m$, ${p}^{\mathcal{T}_m}_c=\sum_{i\in[1,m-1]} \alpha^{\mathcal{T}_i}_s P^{\mathcal{T}_i}_s + \alpha^{\mathcal{T}_m}_c P^{\mathcal{T}_m}_c$, where $\alpha^{\mathcal{T}_i}_b$ and $P^{\mathcal{T}_i}_b$ are the $\mathcal{T}_i$-specific attention and prompt at the server ($b=s$) and client $c$ ($b=c$), respectively. The updated client-side ${p}^{\mathcal{T}_m}_c$ will be uploaded to the server and aggregated based on classical FL~\cite{mcmahan2017communication} to obtain server-side prompt ${p}^{\mathcal{T}_m}_s$. However, such naive aggregation becomes inefficient in asynchronous task arrival. When client $c$ is training task ${\mathcal{T}_m}$, the latest task observed by other clients might be task $\mathcal{T}_n$ ($m<n$), and $\mathcal{T}_n$ will be observed by client $c$ later. Hence, to handle this condition, Fed-CPrompt introduces asynchronous prompt learning.

Instead of waiting for updated prompts of the current task $\mathcal{T}_n$ from all clients before aggregation, we allow task-specific prompt aggregation in parallel. In this way, the previously learned prompt at the server ${p}_s^{\mathcal{T}_m}$ can be refined by ${p}^{\mathcal{T}_m}_c$. Moreover, taking advantage of the task arrival heterogeneity, the training of ${p}^{\mathcal{T}_m}_c$ becomes
\begin{equation}\label{eq:fedCODA}
p_c^{\mathcal{T}_m}=\sum_{i= 1}^{m-1} \alpha^{\mathcal{T}_i}_s P^{\mathcal{T}_i}_s +  \alpha_c^{\mathcal{T}_m} P_c^{\mathcal{T}_m} + \sum_{j= {m+1}}^{n}  \alpha^{\mathcal{T}_j}_s P^{\mathcal{T}_j}_s ,   
\end{equation}
where the first and the third terms are task knowledge from the server, which are frozen when training task $\mathcal{T}_m$. It should be mentioned that although the newest task for client $c$ is $\mathcal{T}_m$, the asynchronous task arrival in FCL allows client $c$ to leverage unseen task knowledge to navigate the local training. By incorporating past and future task representations, we increase the capacity of prompts to learn task-specific instructions.

\subsection{C2Loss: Contrastive and Continual Loss}\label{sec: C2L}
To address the data heterogeneity issue while alleviating forgetting in FCL, we introduce a new loss function, contrastive and continual loss (C2Loss), to regularize local training on clients. The goal of C2Loss is mainly twofold. First, C2Loss accommodates disagreements  between clients due to biased local training with heterogeneous data distribution. Second, C2Loss enforces distinct task-specific prompts construction, which facilitates CL to avoid the forgetting effect.
 Specifically, when learning task $\mathcal{T}_m$ at communication round $r$, we have the C2Loss on client $c\in\mathcal{C}$ given by
\begin{align}\label{eq:C2loss}
    \mathcal{L}_{C2L}&(P_c^{\mathcal{T}_m}(r))=\max\big(|| P_c^{\mathcal{T}_m}(r)  - P^{\mathcal{T}_m}_s(r-1)||_2  
    \\ \notag - \gamma & \min\{|| P_c^{\mathcal{T}_m}(r)  - P^{\mathcal{T}_{i}}_s ||_2 , i\in[1,n], i\neq m \} + \alpha,  0\big), 
 \end{align}
where the first term within the $\max()$ calculates the change of the current prompt compared to that in the previous round. By restricting this change, C2Loss smooths the local update to achieve the first goal. The second term within the $\max()$ finds the most similar prompt to the current prompt based on the distance between the current and all previous prompts. By increasing this distance, C2Loss enforces the discrimination between task-specific prompts to achieve the second goal. Besides, $\gamma>0$ is the hyperparameter to balance the impact between the first two terms. $\alpha \in [0,1]$ represents a margin value that encourages a separation between the first two terms~\cite{schroff2015facenet}. 
 

\subsection{Overall Training}
In Fed-CPrompt, client $c\in\mathcal{C}$ conducts local training with dataset $\mathcal{D}_c^{\mathcal{T}_m}$ for the current task $\mathcal{T}_m$. As discussed in (\ref{eq:fedCODA}), a prompt is constructed based on attention mechanisms, and thus the learnable prompt parameter for client $c$ is defined by $\mathbf{w}_c^{\mathcal{T}_m}=\{P_c^{\mathcal{T}_m}, K_c^{\mathcal{T}_m}, A_c^{\mathcal{T}_m}\}$ ($K$ and $A$ composite the $\alpha$ in (\ref{eq:fedCODA}), detailed in Appendix~\ref{sec:preliminaries}). By incorporating the C2Loss to the cross-entropy loss, we have the local optimization function of client $c$ by
\begin{align}\label{eq:total_objective}
\min_{\mathbf{w}_c^{\mathcal{T}_m},\phi_c^{\mathcal{T}_m}} \mathcal{L}_{CE}(f_{\phi_c}(x; \theta, \mathbf{w}_c),y) + \lambda ~ \mathcal{L}_{C2L}(\mathbf{w}_c^{\mathcal{T}_m}),
\end{align}
where $\phi_c^{\mathcal{T}_m}$ represents the classifier parameter for task $\mathcal{T}_m$. Note that $\mathbf{w}_c$ and $\phi_c$ concatenate both the frozen previous task parameter and the current task learnable parameter as discussed in (\ref{eq:fedCODA}). Besides, $\theta$ is the frozen pretrained model parameters; $(x,y)\in\mathcal{D}_c^{\mathcal{T}_m}$; $\lambda \in [0,1]$ is the hyperparameter balancing losses. 
Both prompt parameters $\mathbf{w}_c^{\mathcal{T}_m}$ and classifier parameters $\phi_c^{\mathcal{T}_m}$ will be uploaded to the server. The server handles asynchronous task arrival by conducting parallel aggregation following classical FL~\cite{mcmahan2017communication}. 
The overall training is illustrated in Algorithm~\ref{alg: overallframework} of Appendix ~\ref{sec: Algorithm}.


\section{Experiments}

\subsection{Experimental Setup}
We evaluate the proposed Fed-CPrompt based on the CIFAR-100 dataset~\cite{krizhevsky2009learning}, a widely used dataset in continual learning for classification tasks. We consider a total of 10 clients in FCL. The server-side knowledge aggregation is based on FedAvg~\cite{mcmahan2017communication}. The evaluation metrics include average accuracy and average forgetting, which are standard metrics used in previous CL research~\cite{wang2023federated,huang2022learn}. To comprehensively evaluate Fed-CPrompt, we consider baseline approaches, including  rehearsal-free FL approaches (i.e., Fed-EWC and Fed-LWF) and recent prompt-based CL approaches (i.e., Fed-CODAP, Fed-DualP, Fed-L2P). Further details on the dataset setup, FL settings, evaluation metrics, and baseline approaches can be found in Appendix~\ref{app: ExperimentSetup}.

\subsection{Experimental Results}

\begin{table}[t]
\small
\centering
\caption{Performance comparison of rehearsal-free FCL methods under iid settings.}
\label{tab:siid_perf}
\begin{tabular}{lrr}
\toprule
Model       & Accuracy & Forgetting \\
\midrule
Fed-EWC     & 9.79           & 84.85  \\
Fed-LwF     & 60.92          & 33.75  \\
Fed-L2P     & 73.19          & 9.80    \\
Fed-DualP   & 76.00          & 9.84   \\
Fed-CODAP   & 77.28          & 6.42   \\
Fed-CPrompt & \textbf{79.43}          &\textbf{4.75}  \\
\bottomrule
\end{tabular}
\vspace{-0.5em}
\end{table}


\begin{table}[t]
\small
\centering
\caption{Performance comparison of prompt-based FCL methods under different non-iid settings (label and quantity skew). }
\label{tab:sniid_perf}
\begin{tabular}{clrr}
\toprule
& Model       & Accuracy & Forgetting \\ 
\midrule
\multirow{3}{*}{Label Skew} 
&Fed-DualP   & 46.28          & 13.16  \\
&Fed-CODAP   & 54.80           & 11.55  \\
&{Fed-CPrompt} & \textbf{65.45} & \textbf{9.15} \\ \midrule 
\multirow{3}{*}{Quantity Skew} 
& Fed-DualP   & 77.57          & 8.38   \\
& Fed-CODAP   & 78.56          & \textbf{5.00}  \\
& Fed-CPrompt & \textbf{81.12}         & 7.75   \\ \bottomrule
\end{tabular}
\vspace{-0.5em}

\end{table}

\textbf{Effectiveness of Fed-CPrompt.} 
We evaluate the effectiveness of Fed-CPrompt under iid and non-iid FL settings. We report the average test accuracy and forgetting over all ten tasks. As illustrated in Table~\ref{tab:siid_perf}, Fed-CPrompt gains a significant performance improvement over all rehearsal-free FCL methods under iid settings. Compared with the best of existing works, Fed-CPrompt achieves around a 2\% increase in Top-1 accuracy and around a 2\% drop in Forgetting. 
It should be mentioned that non-prompt-based methods (Fed-EWC and Fed-LwF) optimize about 86 million parameters, while Fed-CPrompt optimizes only 4 million $(\approx 4.18\%)$ to achieve better performance. Moreover, Fed-CPrompt has better convergence, which significantly reduces the communication cost for FCL.

Besides, we further compare the Fed-CPrompt with other prompt-based FCL baselines under non-iid settings. In the experiments following~\cite{li2020federated}, we consider two non-iid settings: label skew and quantity skew. As illustrated in Table~\ref{tab:sniid_perf}, Fed-CPrompt outperforms the existing prompt-based methods under non-iid settings. In particular, under a challenging label-skew setting, Fed-CPrompt achieves a significant performance improvement by $10.65\%$. 

\textbf{Impact of Asynchronous Continual Learning Tasks.} We demonstrate the effectiveness of Fed-CPrompt under asynchronous task arrival, where the clients train the models on different tasks at the same time. 
As illustrated in Table \ref{tab:asyciid_perf}, the average test accuracy of Fed-CPrompt significantly outperforms the existing methods by $2.07\%$ and $11.79\%$ under iid and non-iid settings, respectively. 
Our findings suggest jointly considering past and future task information can improve the training efficiency of FCL. 
It should be noted that the forgetting of Fed-CPrompt is comparable to or higher than the existing works; however this is due to the high accuracy gained by Fed-CPrompt on the first task. Besides the impact of high accuracy on the first task, we can still observe the substantial advantage of Fed-CPrompt in mitigating catastrophic forgetting, as the average accuracy on all ten tasks achieved by Fed-CPrompt is much higher than the existing works.

\textbf{Impact of C2Loss.} We further perform ablation studies to evaluate the effectiveness of the proposed C2Loss. We compare the performance between FedProx and Fed-CPrompt with and without C2Loss. As shown in Table~\ref{tab:ablation_exp}, Fed-CPrompt with C2Loss achieves the highest accuracy and lowest forgetting compared with the rest two methods. 
This is mainly due to that C2Loss handles inter-task and inter-client knowledge transfer, thereby leading to better task discrimination and improved accuracy.

\begin{table}
\small
\centering
\caption{Performance comparison of asynchronous continual learning under iid and non-iid (label skew) settings.}
\label{tab:asyciid_perf}
\begin{tabular}{clrr}
\toprule
\multicolumn{2}{c}{Model} & Accuracy & Forgetting \\
\midrule
\multirow{3}{*}{iid} 
& Fed-DualP & 70.27 & 17.36 \\
& Fed-CODAP & 73.63 & 10.94 \\
& {Fed-CPrompt} & \textbf{75.70} &  \textbf{9.82} \\
\midrule
\multirow{3}{*}{\shortstack[l]{non-iid}} 
& Fed-DualP & 46.80 &\textbf{ 3.89}\\
& Fed-CODAP & 50.81  & 8.05  \\
& Fed-CPrompt &  \textbf{62.60} & 10.55 \\
\bottomrule
\end{tabular}\vspace{-0.5em}
\end{table}

\begin{table}
\small
\centering
\caption{ Impact of C2Loss under the non-iid (label skew) setting.}
\label{tab:ablation_exp}
\begin{tabular}{lrr}
\toprule
Model       & Accuracy & Forgetting \\ 
\midrule
$\mathcal{L}_{CE}$ w/ FedProx  & 50.81 & 9.16\\
$\mathcal{L}_{CE}$ w/o $\mathcal{L}_{C2L}$ &  71.30 &  13.52 \\
$\mathcal{L}_{CE}$ w/ $\mathcal{L}_{C2L}$  &  \textbf{79.30}& \textbf{ 4.50} \\
\bottomrule
\end{tabular}\vspace{-0.5em}
\end{table}

\section{Conclusion}
This paper proposed Fed-CPrompt, an innovative rehearsal-free FCL framework to alleviate catastrophic forgetting over incremental tasks and facilitate knowledge transfer among distributed and heterogeneous clients. Fed-CPrompt introduces two key components: asynchronous prompt learning to handle asynchronous arrival, and a simple yet effective contrastive continual loss that optimizes prompt parameters while providing additional supervision for learning distinct task-specific prompts. Extensive experiments demonstrate the effectiveness of our proposal.
\section*{Acknowledgment}
The authors thank all anonymous reviewers for their insightful feedback. This work was supported by the National Science Foundation under Grants CCF-2106754, CCF-2221741, CCF-2153381, and CCF-2151238. The work of Miao Pan was supported in part by the US National Science Foundation under Grants CNS-2107057 and CNS-2318664. 

\bibliography{Chapters/papers}


\bibliographystyle{icml2023}

\newpage
\appendix
\onecolumn
\section{Preliminaries for Prompt-based Continual Learning}\label{sec:preliminaries}

In this work, we build upon the technical foundations of prompt-based methods from the prior centralized continual learning research~\cite{smith2022coda,wang2022dualprompt,wang2022learning} to introduce prompts, which can collaboratively learn in heterogeneous federated settings. 
As done in CODA-P, prompt parameters are attached to several multi-head self-attention (MSA) layers in a pre-trained ViT. Define a task-specific prompt parameter for task $\mathcal{T}_m$ as $P^{\mathcal{T}_m} \in \mathbb{R}^{L_P \times D \times \mathcal{T}_m}$, where $L_P$, $D$, and ${\mathcal{T}_m}$ are the prompt lengths, embedding dimension, and the number of prompts for each task, respectively. We consider prefix-tuning to attach prompts to the keys and values of an MSA layer with input  $h\in\mathbb{R}^{L \times D}$ and the query, key, and value as $h_Q, h_K$, and $h_V$. A prompt $p$ is split into $\{P_K, P_V\} \in  \mathbb{R}^{\frac{L_p}{2} \times D}$, which are respectively attached to the key and the value of this layer,  i.e., $MSA(h_Q, [P_K; h_K], [P_V;h_V])$. where $[\cdot;\cdot]$ is a concatenation operation. Since CODA-P achieves SOTA centralized continual learning performance, we adopt the weighted prompt for local training, and the prompt for task $\mathcal{T}_m$ can be calculated by
\begin{equation}\label{CODA_P}
{p}^{\mathcal{T}_m}=\sum_{i\in[1,m]} \alpha^{\mathcal{T}_i} P^{\mathcal{T}_i}, 
\end{equation}
where $P^{\mathcal{T}_m}$ is  a learnable prompt to the current task $\mathcal{T}_m$, $\alpha^{\mathcal{T}_m}=\gamma(q(x)\odot A^{\mathcal{T}_m}, K^{\mathcal{T}_m})$ measures the cosine similarity $\gamma$ between the attended query and the key, where the attended query defined by the element-wise product $\odot$ between query and learnable attention parameter. The query is produced as $q(x)\in\mathbb{R}^D=f(x; {\theta})$, where $f(\cdot;{\theta})$ is the encoder of the pre-trained ViT\footnote{We refer the reader to sections 4.1 and 4.2 of the CODA-p~\cite{smith2022coda} paper for more details.}. For training task $\mathcal{T}_m$, the learnable parameters include $P^{\mathcal{T}_m}$, $K^{\mathcal{T}_m}$, $A^{\mathcal{T}_m}$, and the classification head $\phi^{\mathcal{T}_m}$, whereas  $( \alpha^{\mathcal{T}_i}, P^{\mathcal{T}_i})~\forall i \in[1, m-1]$ is frozen but contributes to the training as in Equation~\eqref{CODA_P}. In addition, the classification head of the previous task $\mathcal{T}_1,\cdots,  \mathcal{T}_{m-1}$ , \textit{i.e} are frozen.

\section{Related Work}\label{sec:RelatedWork}
\textbf{Federated Continual Learning (FCL).} 
FCL performs addresses catastrophic forgetting across multiple clients trained on their private sequential tasks, where a global model is obtained by exchanging task-specific knowledge via a global server. The mainstream FCL research can be roughly divided into two categories: rehearsal-based and rehearsal-free FCL. 
The rehearsal-based research stores and replays information from previous tasks to mitigate the global model's forgetting over time~\cite{dong2022federated,huang2022learn,zizzo2022federated,wang2023federated}. For example, Huang \textit{et al.} proposed FCCL to address the heterogeneity and catastrophic forgetting in federated learning based on buffered data for intra- and inter-domain knowledge distillation~\cite {huang2022learn}. Similarly, Zizzo \textit{et al.} and Wang \textit{et al.} leveraged replay buffers and novel data-sharing approaches based on differential privacy to mitigate forgetting~\cite{zizzo2022federated,wang2023federated}. To tackle the global model's forgetting brought by heterogeneous clients, Dong \textit{et al.} introduced a proxy server to store and select the best old models to assist clients' local training~\cite{dong2022federated}. While successful, the above rehearsal-based FCL research requires large storage space and complex data-sharing strategies to replay past information, making it challenging to scale over time. 

Another category of FCL research is rehearsal-free approaches without storing past information. 
One group of rehearsal-free continual learning (CL) expands the model architecture when encountering new tasks~\cite{rusu2016progressive,li2019learn}. However, most architecture-based approaches require task identity to condition the network during inference, leading to their ineffectiveness for class-incremental or task-agnostic CL scenarios, i.e., the task identity is unknown. In this work, we focus on the practical but more challenging class-incremental FCL in a rehearsal-free manner. 
Existing FCL research along this line proposed regularizing the model with respect to the previous task knowledge when training a new task. For example, Shoham \textit{et al.} and Yoon \textit{et al.} leveraged the weight consolidation method to restrict the updates of the important parameters regarding previous tasks while improving the training performance for the new task~\cite{shoham2019overcoming,yoon2021federated}. Similarly, several recent works implemented knowledge distillation methods to transfer knowledge from the model for the old task to that for the current task~\cite{casado2022concept, usmanova2021distillation, usmanova2022federated,ma2022continual}. 
In addition, ~\cite{shenaj2023asynchronous} investigates the asynchronous-task FCL while using representation loss and a modified aggregation strategy to address the forgetting across multiple clients asynchronously learning respective tasks.
While the aforementioned research enables class-incremental FCL without the rehearsal buffer, they rely on optimizing the entire model on the client side, leading to heavy communication overhead when iteratively exchanging distributed client knowledge in FL, especially for CL scenarios. To address the limitations of existing research, in this work, we propose a novel rehearsal-free FCL approach for class-incremental learning problems based on prompt learning techniques.

\vspace{0.5em}
\textbf{Prompt Learning.} Prompt learning has been a popular transfer learning approach that modifies the input sample with input embedding called prompts, aiming to provide additional information to condition the model to perform downstream tasks~\cite{brown2020language, jiang2020can, shin2020autoprompt}. However, designing the prompt function for various downstream tasks is challenging. Recent research has introduced "soft prompts" to automatically train the learnable prompt parameters to replace the heuristic manual selection, such as the prompt tuning, p-tuning, and prefix tuning~\cite{lester2021power, liu2022p, li2021prefix}. Prompt learning has shown great potential for parameter-efficient transfer learning with a small set of prompt parameters. Taking advantage of the small parameter size, Zhao \textit{et al.}~\cite{zhao2022reduce} and Guo \textit{et al.}~\cite{guo2022promptfl} adopted prompt learning to improve federated learning efficiency. 

Some recent works have implemented prompt learning techniques in CL. Wang \textit{et al.} proposed L2P by using the key-query-based similarity method to select prompts from a prompt pool to instruct different tasks in CL~\cite{wang2022learning}. Later, DualPrompt was introduced as the follow-up to L2P with better CL performance, which learns two sets of disjoint prompt spaces to encode task-specific and task-invariant instructions, respectively~\cite{wang2022dualprompt}. More recently, CODA-Prompt was proposed using an attention-based end-to-end key-query method, which produces the input-conditioned prompts to further improve CL performance~\cite{smith2022coda}. Nevertheless, the above prompt-based CL approaches designed for centralized datasets cannot be directly used for federated learning scenarios, as they ignore the unique challenges raised by distributed nature of clients, such as the heterogeneous data distribution and asynchronous task arrival over clients. To the best of our knowledge, none of the existing prompt learning research has been done for FCL.

\section{Algorithm}\label{sec: Algorithm}
The overall training process includes four main steps (a-d) as shown in Algorithm~\ref{alg: overallframework}.
(a) The server distributes the prompts and model to each new participating device. (b) Each user first freezes the previous prompt parameters. (c) Each user optimizes the local prompt parameters and classifier head following CE loss and Equation ~\eqref{eq:C2loss}.  (d) The clients return the locally trained model to the server. Further,  the server aggregates the model following classical FedAvg~\cite{mcmahan2017communication}. Algorithm~\ref{alg: overallframework} follows steps (a) - (d) until convergence.

\begin{algorithm}[!h]
\caption{Fed-CPrompt}\label{alg: overallframework}
\begin{algorithmic}
\STATE \textbf{Input}: Set of $\mathcal{C}$ clients,  $R$ communication rounds, total tasks $\mathcal{T}$, trainable parameters $\mathbf{w}_s= \{P_s , A_s, K_s\}, \phi_s$, frozen  pretrained large model parameters $\theta$, local epochs $E$, learning rate $\eta$ 
\STATE \textbf{Output}: $\mathbf{w}, \phi$
\end{algorithmic}
\vspace{2pt}

\begin{algorithmic}[1]
\STATE \textbf{Server executes}:
    \STATE  Send the pretrained model $\phi,\theta, \mathbf{w}_s$ parameters to clients.  \hfill \textbf{(a)}

    \FOR {$r \in \{1, \cdots, R \}$}
        \FOR {each client $c \in \mathcal{C}$}
        \STATE $\mathbf{w}_c \leftarrow \ \mathbf{w}_s$
        \STATE $\phi_c \leftarrow \phi_s$
        \STATE $\mathbf{w_c}, \phi_c  \leftarrow $ \textbf{Client Update}($\mathbf{w_c}, \phi_c$)
        
        \ENDFOR
        \STATE Aggregate $\mathbf{w}_c^{\mathcal{T}_m}$ and $\phi_c^{\mathcal{T}_m}$ using FedAvg  \hfill \textbf{(d)}

    \ENDFOR
    \STATE  \textbf{return} $\theta_{r+1}$
\end{algorithmic}
\vspace{2pt}
\begin{algorithmic}[1]
\STATE \textbf{Client Update $(\mathbf{w_c}, \phi)$}:
        \FOR {$i \in [1, n]$}
        \IF{$m \neq n $}
        \STATE Freeze $w_c^{\mathcal{T}_i}, \phi_c^{\mathcal{T}_i}$ \hfill  \textbf{(b)}
        \ENDIF
        \ENDFOR
        \FOR {$e \in E$}
        \STATE $\mathcal{L}_{CE} \leftarrow$ Calculate \textit{cross entropy} loss $\mathcal{L}_{CE} $
        \STATE $\mathcal{L}_{C2L} \leftarrow$ Calculate \textit{C2Loss} using \eqref{eq:C2loss}
        \STATE Update $\mathbf{w}_c^{\mathcal{T}_m}, \phi_c^{\mathcal{T}_m}$   \hfill\textbf{(c)}
        \ENDFOR
    \STATE  \textbf{return} $\mathbf{w}_c^{\mathcal{T}_m}, \phi_c^{\mathcal{T}_m}$
\end{algorithmic}

\end{algorithm}


\section{Implementation details} \label{app: ExperimentSetup}
In this section, we conduct extensive experiments to evaluate the proposed Fed-CPrompt. We first introduce the experimental setup, followed by the experimental results.
Additionally, the same random seed is used to conduct all experiments for reproducibility. 


\vspace{0.3em}
\subsection{Dataset Setup.} \label{sec:datasetup}
The CIFAR-100 dataset consists of $100$ classes with $600$ samples per class. In the experiments, we divide the dataset into $10$ disjoint tasks with $10$ classes per task ($5,000$ training samples per task). We divide the samples on each task among clients following a uniform distribution for the iid settings in federated learning. We implement label-based and  quantity-based distribution skew (i.e., label skew and quantity skew) for non-iid settings with non-iid degree $\beta = 0.5$ following ~\cite{li2022federated}. The test dataset consists of $10,000$ samples, with $1,000$ samples per task. 

\subsection{Federated Learning Settings.} \label{app:fedsett}
The learning rate is set to $lr = 0.0001$. We also deploy an early-stopping mechanism in each task using a validation set.
We consider $\mathcal{C} = 10$ clients, $R=40$ communication rounds, and local epochs $l_{epochs}=5$. The network parameters are optimized using Adam optimizer and a batch size of $128$ images. We split the CIFAR100 dataset into $10$ tasks, each with $10$ classes. This is distributed among the $10$ clients following \S ~\ref{sec:datasetup}.

\vspace{0.3em}
\subsection{Asynchronous Tasks.} We consider an asynchronous scenario where different clients learn from different tasks at the same time. Specifically, we select a random set of 5 clients to participate in the following task $\mathcal{T}_n = \mathcal{T}_{m+1}$, while the remaining 5 clients remain on task $\mathcal{T}_m$. 

\vspace{0.3em}
\subsection{Baseline Approaches.}\label{app:baseline}
We compare our proposed Fed-CPrompt with CODA-Prompt ~\cite{smith2022coda}, Dual prompt ~\cite{wang2022dualprompt}, L2P ~\cite{wang2022learning} applied to federated settings. These prompt-based methods have shown potential parameter-efficient SOTA solutions in continual learning.
Additionally, we consider conventional non-prompt-based rehearsal-free methods to demonstrate the advantages of prompt-based methods.  Specifically, Fed-EWC \cite{shoham2019overcoming} and Fed-LWF ~\cite{usmanova2022federated} provide a fair representation of conventional non-prompt-based rehearsal-free methods in continual learning ~\cite{yoon2021federated, casado2022concept, usmanova2021distillation}. This comparison allows us to show the potential of a prompt-based approach to other rehearsal-free methods. 
The same set of hyper-parameters, such as the learning rate, batch size, and number of rounds is adopted in the baselines and our proposed Fed-CPrompt.

\vspace{0.3em}
\subsection{Prompt parameters.}\label{app:promptpara}
We use prefix-tuning~\cite{li2021prefix} to attach the prompts to layers (1-5) of the pretrained ViT network~\cite{vaswani2017attention}. The total prompt size is $n = 100$, and $10$ prompts per task. Each prompt is set with length $L_p = 8$ and embedding dimension $D = 768$.

\vspace{0.3em}
\subsection{Evaluation metrics.} We evaluate our model on the standard continual learning metrics, including average accuracy and average forgetting, which are widely used in previous works~\cite{li2021prefix,usmanova2022federated,wang2023federated}.  We follow the standard definition of accuracy and forgetting mentioned in~\cite{smith2022coda}. 

\section{Additional Results}
\begin{figure}[!tb]
    \centering

    \subfigure[Total Test Accuracy.]{%
        \includegraphics[width=0.48\textwidth]{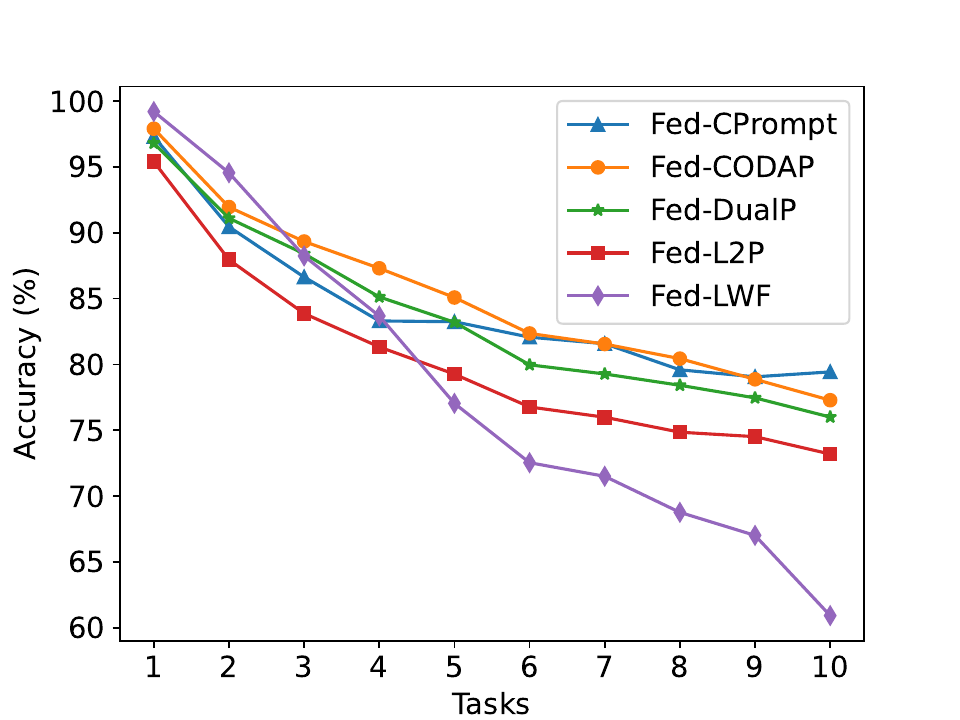}
        \label{fig:sync_acc}
    }
    \hfill
    \subfigure[Overall Forgetting.]{%
        \includegraphics[width=0.48\textwidth]{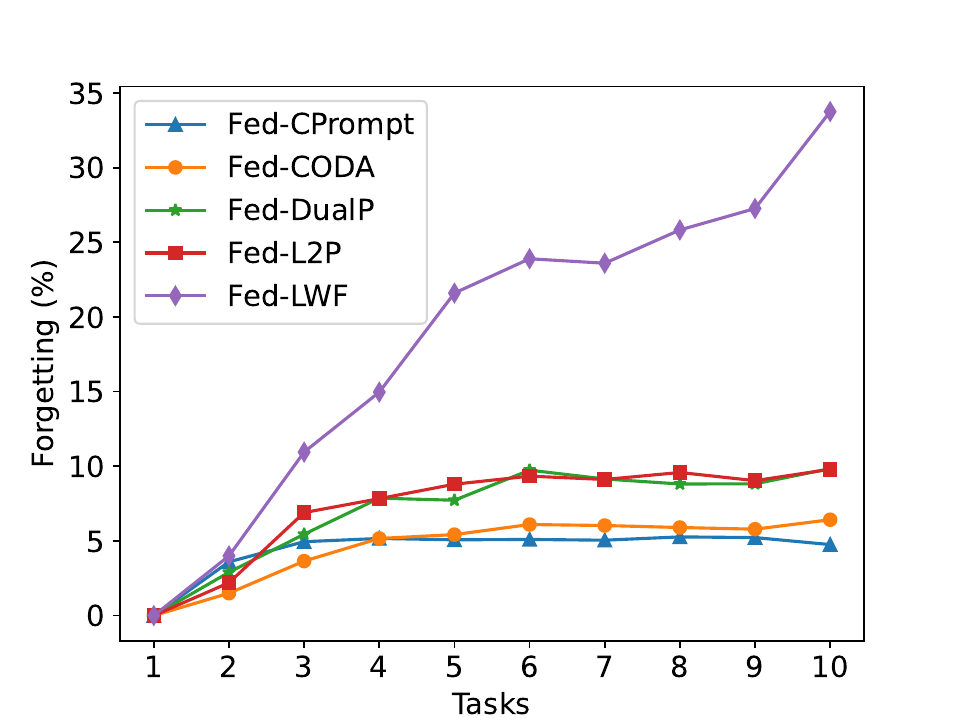}
        \label{fig:sync_forgetting}
    }

    \caption{Total test accuracy and forgetting of the global model after training each incremental task in standard iid settings. Note that the total test accuracy is the average test accuracy of the current task and all the previous tasks.}
    \label{fig:iid_sync}
\end{figure}

\textbf{Training Efficiency.}
Figure \ref{fig:sync_acc} and Figure \ref{fig:sync_forgetting} demonstrate the effect of catastrophic forgetting when training new incremental tasks. Overall, we observe that in Fed-CPrompt retains knowledge from previous tasks, mitigating the catastrophic forgetting issues. Additionally, the accuracy per task is higher due to the increased capacity of prompts compared to prompts used in Dual Prompt and L2P.  
Overall, our findings suggest that prompt-based algorithms, especially Fed-CPrompt, can effectively mitigate the problem of catastrophic forgetting and improve the training efficiency of lifelong learning systems. 

\begin{figure}[!tb]
    \centering

    \subfigure[Total Test Accuracy.]{%
        \includegraphics[width=0.48\textwidth]{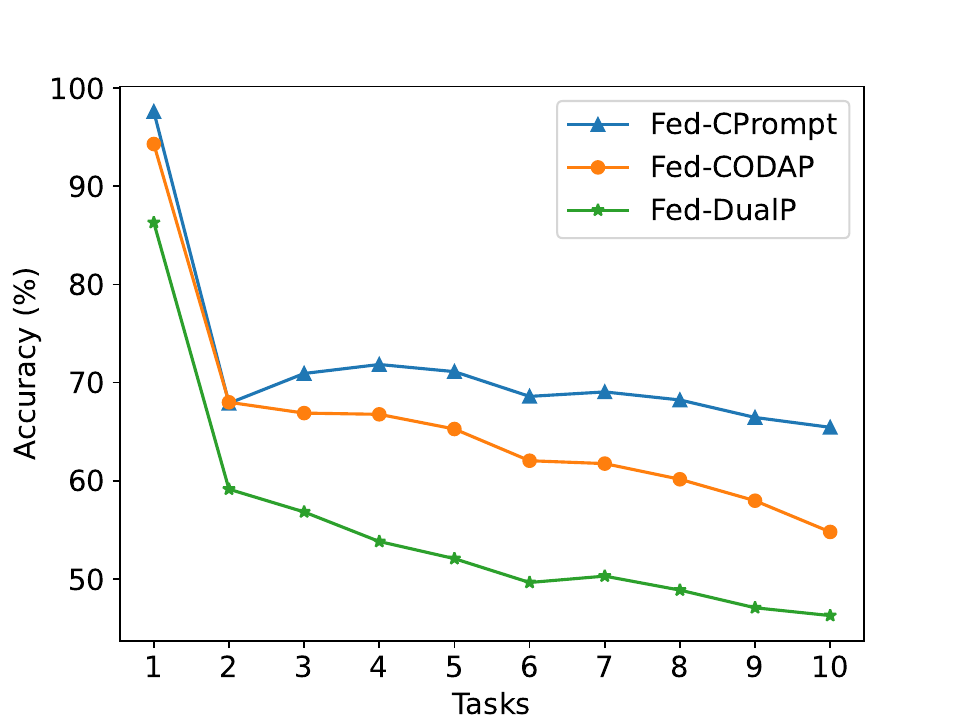}
        \label{fig:sync_niid_acc}
    }
    \hfill
    \subfigure[Forgetting.]{%
        \includegraphics[width=0.48\textwidth]{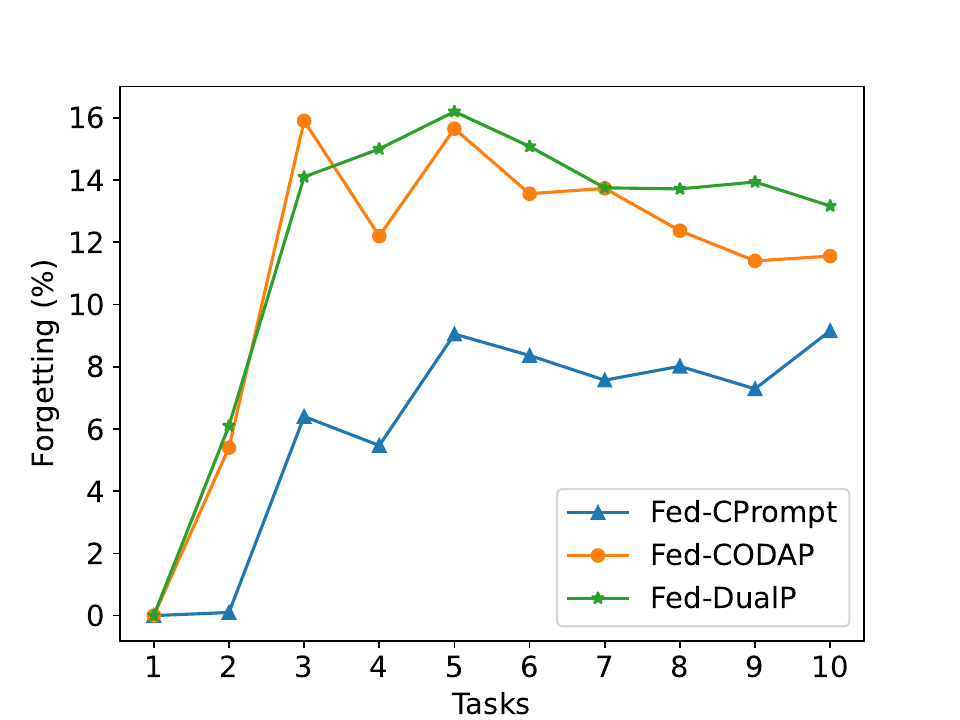}
        \label{fig:sync_niid_forgetting}
    }

    \caption{Total test accuracy and forgetting after training each incremental task on non-iid~(label-skew).}
    \label{fig:niid_sync}
\end{figure}

\begin{figure}[!tb]
    \centering
    \includegraphics[width=\textwidth]{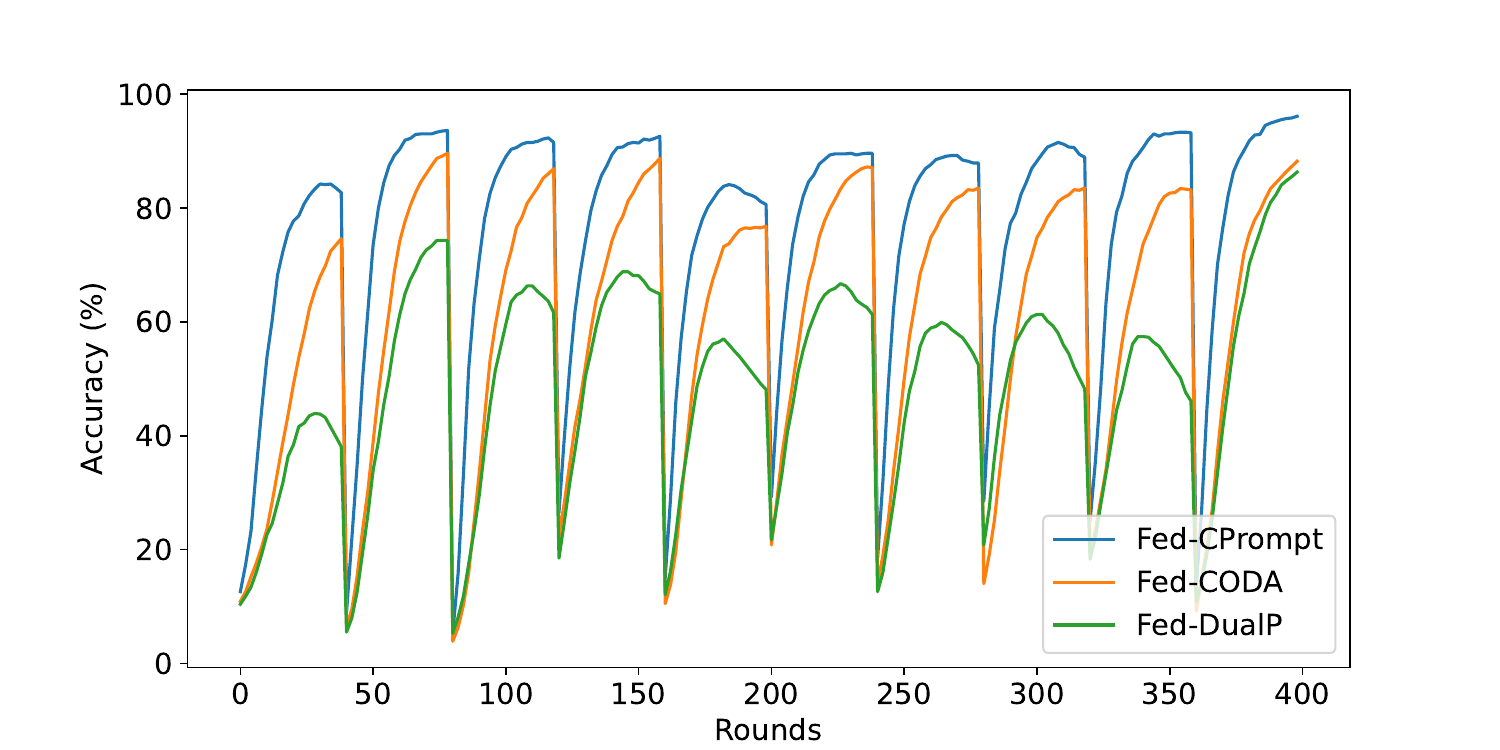}
    \caption{Test accuracy of the prompt-based methods at each round of federated training.  }
    \label{fig:acc_incremental}
\end{figure}

\begin{figure}[!tb]
    \centering
    \includegraphics[width=0.5\textwidth]{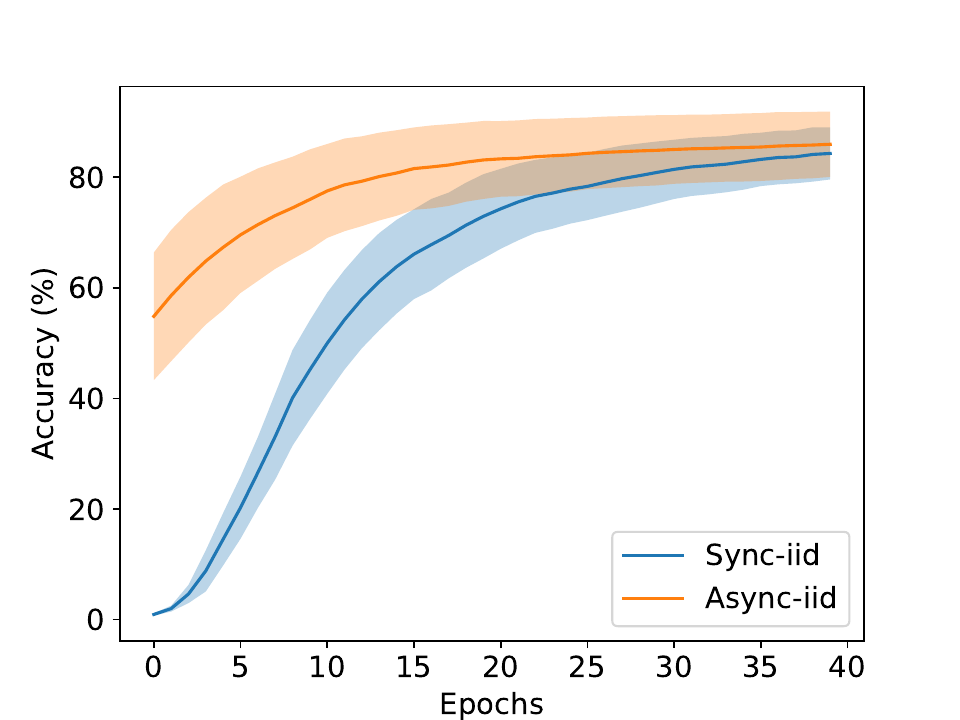}
    \caption{Test accuracy of first 5 tasks in synchronous (sync) and asynchronous (async) iid settings. Under asynchronous settings, prompts in our approach leverage information from other tasks, thus increasing the convergence speed and better initialization.}
    \label{fig:Compare_async_sync}
\end{figure}




\vspace{0.3em}
\textbf{Impact of Asynchronous Continual Learning Tasks.} To investigate the impact of client pacing on the training efficiency of our lifelong learning system, we conduct experiments with varying degrees of client pacing. Specifically, we compare the system's performance when all clients move to the next task simultaneously versus when some clients move to the next task while others are still at the current task. Our results show that when some clients move to the next task, the knowledge of the next task can benefit the current task prompt by providing additional context and improving the convergence speed (Figure~\ref{fig:Compare_async_sync}). The model can leverage the knowledge learned from the next task to understand the current task better, leading to faster convergence and improved accuracy.

Moreover, we also explore the idea of leveraging the prompts from other tasks for example in our design, the clients on task $\mathcal{T}_{m+1}$ leverage prompts from task $\mathcal{T}_m$ to improve the convergence speed of the current task. In addition, clients on task $\mathcal{T_m}$ have an increased capacity due to the additional prompt from task $\mathcal{T}_{m+1}$. Our experiments show that incorporating prompts from previous tasks into the current task prompt can significantly improve the convergence speed and reduce the training time, as shown in Figure \ref{fig:Compare_async_sync}. This is because the model can reuse the knowledge learned from previous tasks and incorporate it into the current task prompt to improve its understanding of the new task.


\end{document}